# Belief-Rule-Based Expert Systems for Evaluation of E-Government: A Case Study

**Mohammad Shahadat Hossain[a], Pär-Ola Zander[b], Md. Sarwar Kamal[a] and Linkon Chowdhury[a]**

[a]*Department of Computer Science and Engineering, University of Chittagong, Chittagong-4331, Bangladesh*

[b]*ICT4D, Aalborg University, Nyhavngsgade 14, 9000 Aalborg, Denmark, E-mail address: poz@hum.aau.dk*

*Corresponding Author*: Mohammad Shahadat Hossain

*Department of Computer Science and Engineering, University of Chittagong, Chittagong-4331, Bangladesh*

*E-mail address*: hossain_ms@cu.ac.bd

mobile: +8801819336660, Tel: +88031637971

## ABSTRACT

E-government development is often complex, with multiple stakeholders, large user bases and complex goals. Consequently, even experts have difficulties in evaluating these systems, especially in an integrated and comprehensive way, as well as on an aggregate level, and thus, there is currently little knowledge about the actual impact and results of e-government. Expert systems are a candidate solution to evaluate such complex e-government systems. However, it is difficult for expert systems to cope with uncertain evaluation data that are vague, inconsistent, highly subjective or in other ways challenging to formalize. This paper presents an approach that can handle uncertainty in e-government evaluation: The combination of Belief Rule Base (BRB) knowledge representation and Evidential Reasoning (ER). This approach is illustrated with a concrete prototype, known as the Belief Rule Based Expert System (BRBES) and implemented in the local e-government of Bangladesh. The results have been compared with a recently developed method of evaluating e-government, showing that the BRBES approach is more accurate and reliable. The BRBES can be used to identify the factors that need to be improved in e-government projects and can juxtapose different scenarios. Thus, the system can be used to facilitate decision making processes under uncertainty.

# 1. Introduction

Governments in many countries are investing vast resources on e-government, aiming to achieve increased efficiency, new business models and perhaps even a more democratic society. However, it is necessary to evaluate the efficiency and effectiveness of the e-government of a country, because little exists knowledge on the impact and results associated with e-government projects and their capacity for real fundamental transformation of relationships between government, citizens, businesses and employees (Esteves & Joseph, 2008; Luna-Reyes, Gil-Garcia, & Romero, 2012). Since the development of e-government is a continuous process, projects need continuous assessment of their nascent or transactional stages, in order to achieve their aims, and for the stakeholders to make appropriate decisions (Al-Sebie & Irani, 2005; Gupta & Jana, 2003). In the following we will argue that present methods of evaluating e-government fail to take various forms of imprecision and complexity into account, and we will introduce an alternative solution.

E-government is very complex since it involves intricate relationships between technological, organisational, institutional and contextual variables (Helbig, Ramón Gil-García, & Ferro, 2009). These variables play an important role in determining the characteristics variables, such as the quality of user environment, electronic management, e-services etc. (Luna-Reyes et al., 2012). For example, the quality of e-government applications (personalization, usability, accessibility etc.) are related to a series of determinants, such as institutional and organizational frameworks, as well as to the technological infrastructure. High quality applications will produce expected results and benefits, such as transparency and accountability, efficiency and effectiveness, citizen participation, effectiveness and program policy, and ultimately high a quality of public service. The above variables can be grouped into three categories, namely Determinants (D), Characteristics (C) and Results (R); and they are intricately interrelated. Therefore, in order to capture the complexity of e-government, an evaluation model should be developed based on these three categories of variables (Luna-Reyes et al., 2012). This approach would allow the evaluators to perceive how the results are produced and to identify the contributing role of each variable in the overall evaluation of the e-government in an integrated way. Other approaches (Dawes, 2008; Esteves & Joseph, 2008; Gupta & Jana, 2003; Karunasena & Deng, 2012; Raus, Liu, & Kipp, 2010; Stowers, 2004; Verdegem & Verleye, 2009) do not allow such evaluation. We consider that an overall evaluation establishes a measure of the usefulness of the system, cf. (Nielsen, 1993), who defines it as "*Usefulness* is the issue of whether the system can be used to achieve some desired goal." (ibid, p. 24)

Such an approach would allow the decision makers to develop an appropriate policy, enabling the enhancement of future e-government initiatives of a country. It is interesting to note that many of the variables that e-government literature deals with, and which we will delve more into in the following, cannot be measured with precision or with 100% certainty. The reason for this is that most of the variables are subjective in nature, for example, usability, which cannot be measured with 100% certainty. Hence, any approach to evaluate e-government should consider this uncertainty phenomenon.

Since e-government evaluation is a complex issue, algorithmic solutions cannot be considered. A problem of this nature is often handled by developing an expert system. Expert systems have already shown their applicability within important fields of e-government evaluation (Magoutas & Mentzas, 2010; Shan, Xin, Wang, Li, & Li, 2013; H.-H. Yang, Liu, Chang, & Yang, 2012). An expert system consists mainly of two important parts: The knowledge-base and the inference engine. Various knowledge representation languages such as Propositional Logic (PL), First-order Logic (FOL), and Fuzzy Logic (FL) are used to develop the knowledge base (Angulo et al., 2012; Liu, Singonahalli, & Iyer, 1996), and reasoning mechanisms such as forward chaining (FC) and backward chaining (BC) are used to develop inference engines (Russel & Peter, 2009). However, neither PL nor FOL are equipped with schemas that can capture uncertainties. On the other hand, although FL handles some uncertainties due to ambiguity, vagueness or imprecision, it cannot handle other types of uncertainties such as ignorance, incompleteness, and ignorance in fuzziness, which may exist within different categories of variables associated with e-government. Therefore, in order to develop such expert systems, a knowledge representation language is required which will enable the handling of different types of uncertainties that exist within the variables of e-government, as mentioned previously. Furthermore, inference mechanisms such as FC and BC are not equipped to handle uncertainties, and hence, the expert system should have the inference engine, facilitating the handling of uncertainties. For these reasons, the uncertain data that exist in the evaluation of e-government, need to be processed by using a refined knowledge representation schema and inference mechanism. Hence, this article considers the employment of a recently developed Belief Rule Base (BRB) inference methodology using the Evidential Reasoning approach (RIMER) (Jian-Bo Yang, Jun Liu, Jin Wang, How-Sing Sii, & Hong-Wei Wang, 2006; Zhou, Hu, Xu, Yang, & Zhou, 2010) for the design and development of an expert system. RIMER consists of two main parts. BRB is a knowledge representation schema used in the first part, while Evidential Reasoning (ER) is used as an inference mechanism in the second part. BRB is the extended form of traditional IF-Then rule bases, and contains appropriate schemas to capture different types of uncertainties and allow the handling of non-linear causal relationships. Case-Based Reasoning (CBR) is also used as a method to build a knowledge base (Fang et al., 2014; B. Sun, Xu, Pei, & Li, 2003); however it accepts anecdotal evidence as its main operating principle and hence, the deduction of inference based on CBR is unreliable. In contrast, for a number of reasons, rule bases appear to be one of the most common forms for expressing various types of knowledge (Jian-Bo Yang et al., 2006; R. Sun, 1995). As such, rule based expert systems usually constructed from human knowledge in the form of IF-Then rules become the most visible and fastest growing branch of Artificial Intelligence (Jian-Bo Yang et al., 2006; R. Sun, 1995). The capturing of human knowledge is essential to enabling the evaluation of e-government and hence, building a knowledge base by using a rule base should be appropriate than that of CBR. Knowledge base building by the use of a rule base is also common in other areas of a problem domain, such as agricultural ecosystem management (L. Xu, Liang, & Gao, 2008). The Evidential Reasoning approach (M.S. Hossain Md, Salah Uddin Chowdury, & Sarker, 2013; J.-B. Yang & Sen, 1997; Yang, Jian-Bo & Singh, 1994) deals with multiple attribute decision analysis (MADA) of problems with both qualitative and quantitative attributes under uncertainties and hence facilitates the handling of uncertainty in the inference process.

The purpose of the remainder of this article is to outline the development of a Belief Rule Based Expert System (BRBES) that can be applied to evaluate e-government in a country. Bangladesh was chosen as the case study area for the application of the system. It will be demonstrated that the result generated from the system is more accurate than the recently proposed multidimensional model for evaluating e-government (Luna-Reyes et al., 2012). This article is divided into seven sections, including the above introduction. The second section presents the variables associated with Determinants, Characteristics and Results and their associated different types of uncertainties. The third section describes methodology used to develop the expert system. The fourth section presents the design and implementation of the expert system. The application of the BRBES and its validation are presented in section five. Section six presents the discussion, while section seven includes some final comments and suggestions for future research.

## 2. Electronic Government Variables and their Uncertainties

Table 1 illustrates the common e-government variables associated with the three different categories, namely determinant, characteristics and results (ibid). We departed from the LRF and consulted additional literature when further elaboration was needed, as indicated in Table 1 below. The table also shows the uncertainty associated with each variable. An operational definition of each variable has been obtained during visits to various ICT projects in Bangladesh, such as the Union Information Service Centre, and District web portal projects located in the Chittagong, Cox's bazar and Bandarban districts of Bangladesh. This approach of obtaining the operational definition of each variable facilitates the determining of different types of associated uncertainties.

### *2.1. Determinants*

A better understanding of the determinants of the characteristics and the results are essential. The main determinants identified in the literature consist of quality of the information and existing data to feed into the systems, technological infrastructure and compatibility, organisational and management-related characteristics, existing legal and institutional frameworks and potential demand. Again, these variables are subjective in nature and hence contain various types of uncertainties, as noticed during our field visits, as illustrated in Table 1.

### *2.2. Characteristics*

Characteristics of e-government facilitate the evaluation of how the technical and functional requirements were made (Luna-Reyes et al., 2012). Examples of such requirements are usability, quality of information available on web sites and in systems, privacy, security, interaction, integration, personalization, accessibility, and services. The uncertainties associated with the quality of information available on websites and in systems include imprecision, incompleteness and vagueness. Imprecision may exist because of the absence of indication of quality or acceptance of quality. Incompleteness may exist, for instance, because a user does not wish to disclose his opinions on a given matter. Vagueness may exist because information and the interface of a website are usually evaluated in terms of linguistic terms such as good, very good and excellent, which are relatively imprecise. Therefore, the uncertainties associated with the other variables were identified as illustrated in Table 1 during our visits to the case study areas.

**Table 1**

The measurement framework with e-government.

| | **Uncertainty Type** | **Discussion** |
|---|---|---|
| **Determinants** | | |
| Quality of the information and existing data to feed the system | Incompleteness, ignorance | Whether the information in the websites (for example district or union web portal/software) is adequate and of a good quality |
| Technological infrastructure and compatibility | Vagueness | Quality of technical equipment and environment (power supply, and Internet service) |
| Organizational and management-related characteristics | Incompleteness, ignorance | Determines the workflow and performance of an organization |
| Existing legal and institutional framework | Vagueness | How stakeholders maintain accountability and regulations in their ICT services |
| Potential demand | Incompleteness | describes to what extent a fully informed citizen would ask for the services in question |
| ***Characteristics*** | | |
| Quality of information available on websites and in systems | Vagueness | Refers to the arrangement, design and orientation of the system elements or web content with proper guidelines as to how to use this |
| Services | Inconsistency | The services received by the users or people |
| Interaction (Berntzen & Olsen, 2009) | Imprecision, incompleteness and vagueness | Refers to the degree of interaction of the users with the systems/websites |
| Integration (Accenture, 2004) | Incompleteness and risk | The integrated environment, enabling the components of e-government systems to work together |
| Personalization (Accenture, 2004) | Vagueness and incompleteness | The tailoring of services to the individual citizen rather than ‘onesize fits all’ |
| Security (Mary Maureen Brown, 2001) | Vagueness and incompleteness | Security indicates the protection of web content and hence requires good security policies at all levels |
| Privacy (M.M. Brown & Brudney, 2003) | Vagueness and incompleteness | Refers to whether knowledge to control personal information is provided |
| Accessibility (Gant, Gant, & Jhonson, 2002) | Incompleteness | People can easily access the system or websites even if they are suffering from some impairment |
| Usability | Imprecision and incompleteness | Systems have learnability, memorability, safety and satisfactory features |
| ***Results*** | | |
| Statistics on system usage | Incompleteness | Measure the interactions of users on ICT tools or services as percentages or ratios |
| Quality of public services | Imprecision and incompleteness | The degree to which users experience good quality |
| Efficiency and productivity | Vagueness | The system is less prone to errors and has faster processing time |
| Effectiveness of programs and policies | Vagueness | The degree to which systems results are aligned to political programs |
| Transparency and accountability | Imprecision, vagueness and incompleteness | 1) Can the user follow what happens in the workflow of government transactions, or is it black-boxed? And 2) Can a specific stakeholder be held accountable for mistakes (and successes) |
| Citizen participation | Imprecision and incompleteness | The degree to which citizens of all classes can influence the systems |
| Changes in the regulatory framework | Inconsistency | The process which ensures the routine upgrading of the existing system |

### *2.3. Results*

The results manifest the benefits, identified as the impact of e-government. The main e-government result variables identified in the literature are the following: statistics on systems usage, quality of public services, efficiency and productivity, effectiveness of programs and policies, transparency and accountability, citizen participation and changes in the regulatory framework. These variables are subjective in nature, and the various types of uncertainties associated with them were identified during the field visits, as shown in Table 1.

An expert system would generate appropriate results if it encapsulated these three types of variables under uncertainty.

# 3. BRB Methodology for an Expert System to Assess E-Government

A Belief Rule Base (BRB) is a knowledge representation schema which allows the capturing of uncertain information. Evidential Reasoning (ER) is used as the inference methodology in the Belief Rule Based Expert system (Jian-Bo Yang et al., 2006). ER is mainly used to aggregate the rules in the BRB, either in a recursive or an analytical way (Y.-M. Wang, Yang, & Xu, 2006). This approach is widely known as the RIMER methodology. A BRB can capture nonlinear causal relationships under uncertainty between antecedent attributes and the consequent, which is not possible in traditional IF-THEN rules. In e-government evaluation, determinants, characteristics and results can be considered as examples of consequent attributes, while variables under each category can be considered as the antecedent attributes. In e-government evaluation, the causal relationship between each category and its corresponding variables may be non-linear and complex; and different types of uncertainties are associated with each variable, as shown in Table 1.

Therefore, BRB can be considered to be an appropriate knowledge representation schema to build the knowledge base of the expert system. Fig.1 shows the architecture of the Belief Rule Based System, consisting of its input, inference procedures and output components. Inference procedures consist of input transformation, rule activation weight calculation, rule update mechanisms, followed by the aggregation of the rules of a BRB by using ER. This aggregation facilitates the obtainment of the distribution of belief degrees for the consequent attribute for the given values of antecedent attributes of a BRB.

This section presents the BRB, which will provide the understanding of the knowledge acquisition and representation procedures. Eventually, this will be used to build the initial BRB for a problem domain, in our case an e-government evaluation. The procedure of input transformation is introduced in this section. The aggregation procedure of rules in an initial BRB uses the ER methodology, which is considered as the inference mechanisms presented in this section. This will allow the calculation of the belief distribution of a consequent attribute for certain input values of antecedent attributes of a BRB. The procedures of the rule update mechanism are also discussed.

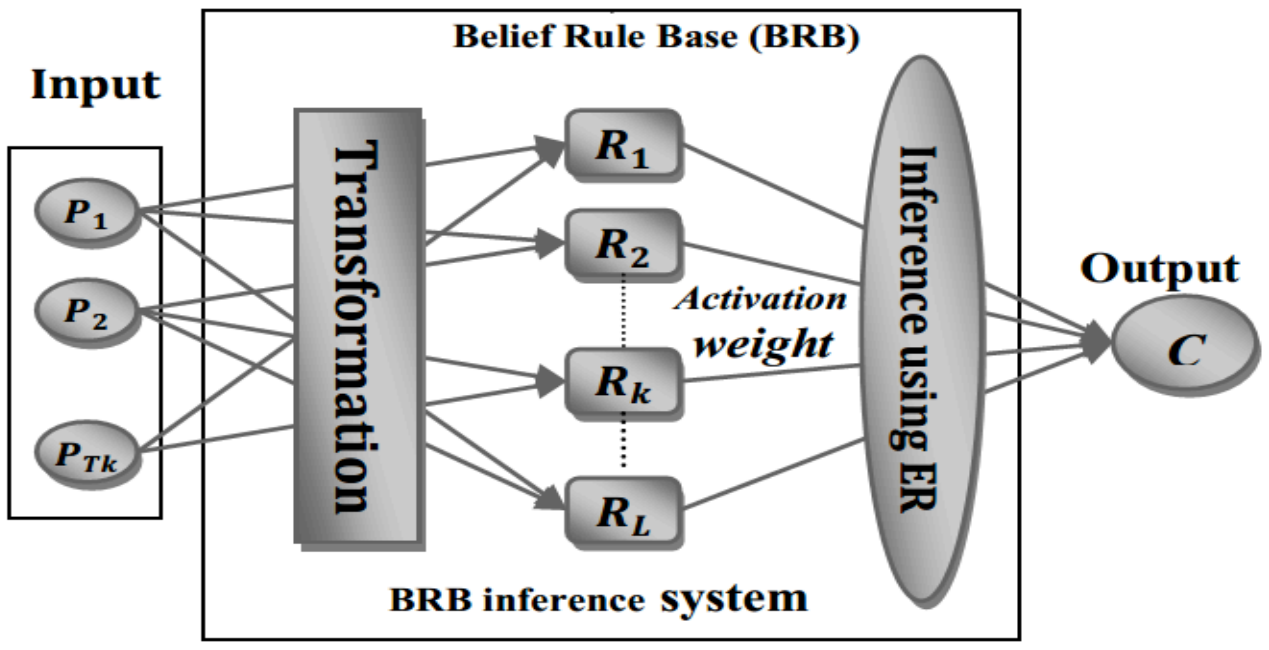

Fig. 1. Single-layer BRB inference architecture with RIMER methodology

*3.1. Domain knowledge representation*

Belief Rules are the key constituents of a BRB; they include belief degrees and are the extended form of the traditional IF-THEN rule. In a belief rule, each antecedent attribute takes on referential values, and each possible consequent is associated with belief degrees. An example of referential values for one of the antecedent attributes or variables in e-government evaluation could be “Excellent”, “Very Good”, “Good”, “Satisfactory” or “Poor”. The knowledge representation parameters are rule weights, antecedent attribute weights and belief degrees in consequent, which are not available in a traditional IF-THEN rule. A belief rule (R) can be defined in the following way:

$$R_k : \begin{cases} IF(P_1 \text{ is } A_1^K) \cap (P_2 \text{ is } A_2^K) \cap (P_3 \text{ is } A_3^K) \cap \dots \dots \dots \cap (P_{T_k} \text{ is } A_{T_k}^K) \\ THEN\{(C_1, \beta_{1k}), (C_2, \beta_{2k}), (C_3, \beta_{3k}), \dots \dots \dots, (C_N, \beta_{Nk})\} \end{cases}$$

$$(\beta_{jk} \geq 0, \sum_{j=1}^{N} \beta_{jk} \leq 1) \text{ with a rule weight } \theta_k, \text{ attribute weights } \partial_{k1}, \partial_{k2}, \partial_{k3}, \dots \dots, \partial_{kT_k} \ k \in \{1, \dots \dots, L\}$$

***P1, P2, P3 …*** represent the antecedent attributes in the ***k***th rule. $\boldsymbol{A_i}$ represents one of the referential values of the ***i***th antecedent attribute $\boldsymbol{P_i}$ in the ***k***th rule. $\boldsymbol{C_j}$ is one of the consequent reference values of the belief rule. $\beta_{jk}$ $j$=1,……,, $k$=1,…… ,$L$ is the degree of belief to which the consequent reference value is believed to be true. If $\sum_{j=1}^{N} \boldsymbol{\beta_{jk}} = \mathbf{1}$, the ***k***th rule is said to be complete; if not, it is incomplete. ***Tk*** is the total number of antecedent attributes used in ***k***th rule. ***L*** is the number of all belief rules in the rule base. ***N*** is the number of all possible referential values of consequent in a rule.

An example of a belief rule in the domain of e-government evaluation can be written in the following way:

*IF Interaction is Excellent AND Integration is Very Good AND Personalization is Satisfactory, THEN User Environment is* {(*Excellent*, 0.00) , (*Very Good*, 0.2222) , (*Good*, 0.7778) ,(*Satisfactory*,0.00) ,( *poor*,0.00)}.. (1)

Where {(Excellent, 0.00), (Very Good, 0.2222), (Good, 0.7778), (Satisfactory, 0.00), (Poor, 0.00)} is the belief distribution associated with the "User Environment" consequent of the belief rule as represented in (1). "Excellent", "Very Good", "Good", "Satisfactory" and "Poor" are the referential values of the consequent "User Environment" of the rule. The belief distribution states that the degree of belief associated with "Excellent" is 0%, 22.22% with "Very Good", 77.78% with "Good", 0% with "Satisfactory" and 0% with "Poor". In this belief rule, the total belief is (0+0.2222+0.7778+0+0) =1 and hence, the assessment is complete.

*3.2. BRB Inference System*

The BRB inference system as illustrated in Fig.1 consists of various components, all aimed at the aggregation of the rules of a BRB by using ER. This section presents a discussion of the central components.

*3.2.1. Input Transformation*

The input transformation of a crisp (non-fuzzy) value of an antecedent attribute $\boldsymbol{P_i}$ consists of distributing the value into belief degrees of different referential values of that antecedent. This is equivalent to transforming an input into a distribution on referential values of an antecedent attribute by using their corresponding belief degrees (Yang et al., 2006). The *ith* value of an antecedent attribute at a specific point in time can likewise be transformed into a distribution across the referential values defined for the attribute by using their belief degrees. The assessment of the input value $\boldsymbol{Ai}$ is shown in the equation below (3).

$H(\boldsymbol{Ai}) = Aij, \alpha ij, j=1,\cdots,ji, i=1,\ldots\ldots,Tk\ldots$ (3)

Here, H is used to show the assessment of the belief degree assigned to the input value of the antecedent attribute. In the above equation, $\boldsymbol{A_{ij}}$ (ith value) is the *j*th referential value of the input, $\boldsymbol{P_i}$. $\boldsymbol{\alpha_{ij}}$is the belief degree to the referential value with $\alpha_{ij} \geq 0.\ \sum_{j=1}^{j_i} \alpha_{ij} \leq 1 (i = 1, \ldots, T_k)$ and $\boldsymbol{j_i}$is the number of the referential values.

In this research, the input value of an antecedent attribute is collected from the people working in the different ICT projects in Bangladesh. They gave the value in a 1-10 scale. This value was then distributed in terms of belief degree of the different referential values [Excellent, Very Good, Good, Satisfactory, Poor] of the antecedent attribute. The referential values $\boldsymbol{A_{ij}}$ can be assigned some preference/utility values $h_{ij}$. For example, the “Excellent” referential value can be assigned utility value *hi5*=“10”, “Very Good” can be assigned *hi4*= “7”, “Good” can be assigned $h_{i3}$ = "6", "Satisfactory" can be assigned $h_{i2}$ ="5", and "Poor" can be assigned $h_{i1}$ = "4". The above procedure of input transformation is elaborated in the equations (4) to (7) below.

$if\ h_{i5} \geq A_i \geq h_{i4},\ then,\ \alpha_{i4} = \frac{h_{i5} - A_i}{h_{i5} - h_{i4}}, \alpha_{i5} = 1 - \alpha_{i4}$ ...(4) $if\ h_{i4} \geq A_i \geq h_{i3},\ then,\ \alpha_{i3} = \frac{h_{i4} - A_i}{h_{i4} - h_{i3}}, \alpha_{i4} = 1 - \alpha_{i3}$ .....(5)

$if\ h_{i3} \geq A_i \geq h_{i2},\ then,\ \alpha_{i2} = \frac{h_{i3} - A_i}{h_{i3} - h_{i2}}, \alpha_{i3} = 1 - \alpha_{i2}$ ...(6) if $h_{i2} \geq A_i \geq h_{i1},\ then\ \alpha_{i1} = \frac{h_{i2} - A_i}{h_{i2} - h_{i1}}\quad \alpha_{i2} = 1 - \alpha_{i1}$ ""(7)

*3.2.2. Calculation of activation* weight

It is important to note that the rule ($k$th) of an initial rule base of a BRB is constructed by taking account of only one of the referential values ($A_{ik}$, which is the element of $j$) of an antecedent attribute ($P_i$). Therefore, it is necessary to determine the degree of belief $\alpha_{ik}$ (which is the element of $\alpha_{ij}$) of this referential value ($A_{ik}$), which can be defined as the matching degree at which the belief is matched. $\alpha_{ij}$ can be calculated by using (4) to (7). When a matching degree is assigned to the referential values of the antecedent attributes of a rule, the rule is considered as activated. This phenomenon is called Packet Antecedent of a rule. When the $k$th rule is activated, the weight of activation of the $k$th rule, $w_k$, is calculated by using the flowing formula (Jian-Bo Yang et al., 2006).

$$\omega_k = \frac{\theta_k \alpha_k}{\sum_{j=1}^{L} \theta_j \alpha_j} = \frac{\theta_k \prod_{i=1}^{T_k} (\alpha_i^k)^{\overline{\delta_{ki}}}}{\sum_{j=1}^{L} \theta_j \left[\prod_{l=1}^{T_k} (\alpha_l^j)^{\delta_{jl}}\right]} \text{......(8) and } \overline{\delta_{ki}} = \frac{\delta_{ki}}{max_{i=1,\dots,T_k}\{\delta_{ki}\}} \dots (8)$$

Where $\delta_{ki}$ is the relative weight of $P_i$ used in the $k$th rule, which is calculated by dividing weight of $P_i$ by maximum weight of all the antecedent attributes of the *kth* rule. By doing so, the value of $\delta_{ki}$ becomes normalized, meaning that the range of its value should be between 0 and 1. $\alpha_k = \prod_{i=1}^{T_k} (\alpha_i^k)^{\overline{\delta_{ki}}}$ $\alpha_k$ is the combined matching degree, which is calculated by using a multiplicative aggregation function.

*3.2.3. Belief Update of Incomplete Data*

After the $\boldsymbol{k}$th rule as shown in (1) is activated, the incompleteness of the consequent of a rule can also result because of a lack of data in the antecedents. While collecting data, we noticed that some people were unable to comment on some of the e-government variables mentioned previously, meaning they have incomplete knowledge or ignore the issue. An incomplete input for an antecedent attribute will lead to an incomplete output in each of the activated rules in which the attribute is considered. The original belief degree $\overline{\beta_{ik}}$ in the $\boldsymbol{i}$th consequent $\boldsymbol{C_i}$ of the $\boldsymbol{k}$th rule is updated based on the actual input information (D.-L. Xu et al., 2007).

$$\boldsymbol{\beta}_{ik} = \overline{\boldsymbol{\beta}_{ik}} \frac{\sum_{t=1}^{T_k}\left(\tau(t,k)\sum_{j=1}^{J_t}\alpha_{tj}\right)}{\sum_{t=1}^{T_k}\tau(t,k)} \text{......(9)}$$

Where $(t,k) = \begin{cases} 1, & if\ P_i\ is\ used\ in\ defining\ R_k(t = 1, \ldots, T_k) \\ 0, & otherwise \end{cases}$

Here $\overline{\beta_{ik}}$ is the original belief degree and $\boldsymbol{\beta ik}$ is the updated belief degree.

*3.2.4. Rule Aggregation*

The ER approach (M. S. Hossain, Chowdhury, & Sarker, 2013; Mahmud & Hossain, 2012; Tang et al., 2012; J.-B. Yang & Singh, 1994) was developed to handle multiple attribute decision analysis (MADA); this is a problem with both qualitative and quantitative attributes under uncertainty. This ER approach is used to aggregate all the packet antecedents of the $\boldsymbol{L}$ rules to obtain the degree of belief of each referential value of the consequent attribute by taking account of given input values $\boldsymbol{P_i}$ of antecedent attributes. Using the analytical ER algorithm (Want et al, 2006), the conclusion *O(Y),* consisting of referential values of the consequent attribute*,* is generated. Equation (10) as given below formalizes this:

$$O(Y) = S(P_i) = \{(C_j, \beta_j),\ j = 1, \dots, N\}\ldots(10)$$

where $\beta_j$ denotes the belief degree associated with one of the consequent reference values such as $C_j$. $\beta_j$ is calculated by analytical format of the ER algorithm as illustrated in equation (4).

$$\beta_j = \frac{\mu \times \left[\prod_{k=1}^{L}\left(\left(\omega_k \beta_{jk} + 1 - \omega_k \sum_{j=1}^{N} \beta_{jk}\right)\right) - \prod_{k=1}^{L}\left(1 - \omega_k \sum_{j=1}^{N} \beta_{jk}\right)\right]}{1 - \mu \times \left[\prod_{k=1}^{L} 1 - \omega_k\right]} \ldots.(11)$$

$$\mu = \left[\sum_{j=1}^{N} \prod_{k=1}^{L}\left(\omega_k \beta_{jk} + 1 - \omega_k \sum_{j=1}^{N} \beta_{jk}\right) - (N-1) \times \prod_{k=1}^{n}\left(1 - \omega_k \sum_{j=1}^{N} \beta_{jk}\right)\right]^{-1}$$

where $\omega_k$ is the activation weight, which also needs to be calculated (see Yang et al, 2006). The final combined result or output generated by ER is represented by $C_1,\beta_1$ , $C_2,\beta_1$ , $C_3,\beta_1$ ,………, $C_N,\beta_N$, where $\beta_j$ is the final belief degree attached to the $j$th referential value $C_j$ of the consequent attribute, obtained after combining all activated rules in the BRB by using ER.

### *3.2.5. Output of the BRB System*

The output of the BRB system is not crisp/numerical value. Hence, this output can be converted into crisp/numerical value by assigning utility score to each referential value of the consequent attribute.

$$H(A^*) = \sum_{j=1}^{N} u(C_j) B_j \ldots\ldots.. (12)$$

where $H(A^*)$ is the expected score, expressed as a numerical value, and $u(C_j)$ is the utility score of each referential value.

## 4. A Belief Rule Based Expert System (BRBES) for Evaluating E-government

The architecture of the BRBES for the evaluation of e-government along with an implementation strategy for this expert system are introduced in this section. It also represents the components of this architecture, including knowledge base, inference engine and system interface. We provide this overview in order to provide the reader with an idea of how comprehensive it is to implement a BRBES. Furthermore, it is difficult to follow the evaluation of the BRBES as an alternative to the LRF evaluation method without at least an idea of its inner mechanics.

*4.1. System architecture*

System architecture represents how its components consisting of input, process and output are organized. It also considers the pattern of the system organization, known as architectural style. The BRBES developed here adopts a three-layer architectural style, including an interface layer (used to obtain antecedent referential value from the people involved with ICT project and to show the system outputs), an application layer (consisting of inference engine with procedures including input transformation, rule activation, rule update and rule aggregation by using ER) and the data management layer (consisting of an initial rule-base developed by using the BRB and survey data collected from the ICT personnel or people using the system). A relational database (MySQL) has been used as back-end to store and manipulate the initial BRB, which represents the knowledge base of the system (see Fig. 2).

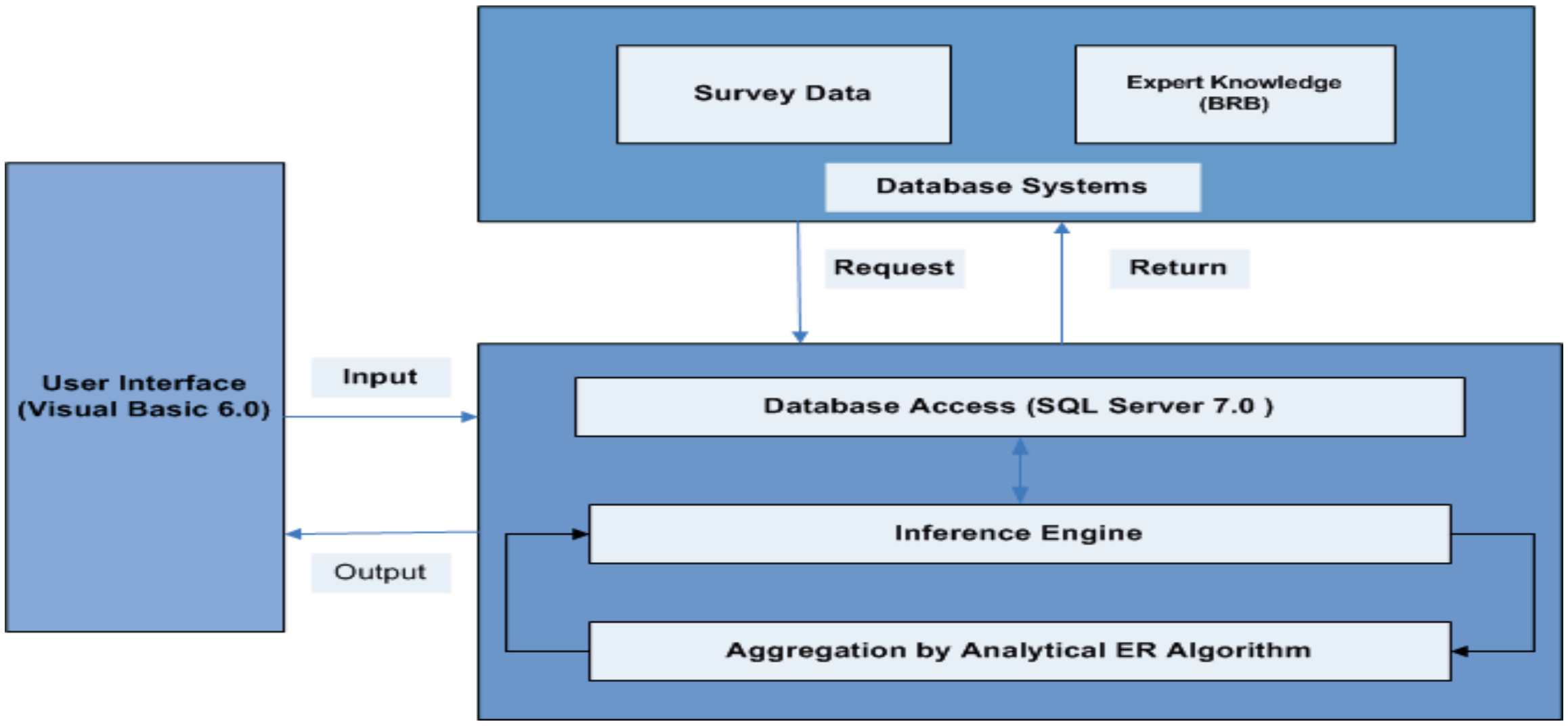

Fig. 2. Architecture of the BRBES

The inference engine works in the following way: i) It first reads input data acquired from the ICT personnel or people involved ii); the input data are then transformed by using (4) to (7), iii) calculates the activation weight of all rules using (8), iv) updates belief degree of consequence in rules using (9) and finally, v) aggregates all rules by using (11).

### *4.2. Knowledge Base Construction in the BRBES*

In order to construct the knowledge base for this expert system, a BRB framework (taking into account variables associated with the determinants, characteristics and results shown in Table 1) was developed, as illustrated in Fig. 3. From this framework, it can be observed that the input variables that determine the evaluation of e-government include all the variables shown in Table 1. However, to reduce the computational complexity of the BRBES, nine variables under the category of 'Characteristics' were again categorized into three sub-groups, i.e. 'user environment' (C1), 'resource management' (C2) and 'authentication protocol' (C3). Fig. 3 illustrates the variables in each sub-group. For the same reason, seven variables under group 'results' were categorized under two sub-groups, i.e. 'result analysis' (R1) and 'result specification' (R2); and the related variables were assigned to sub-groups as shown in Fig. 3.

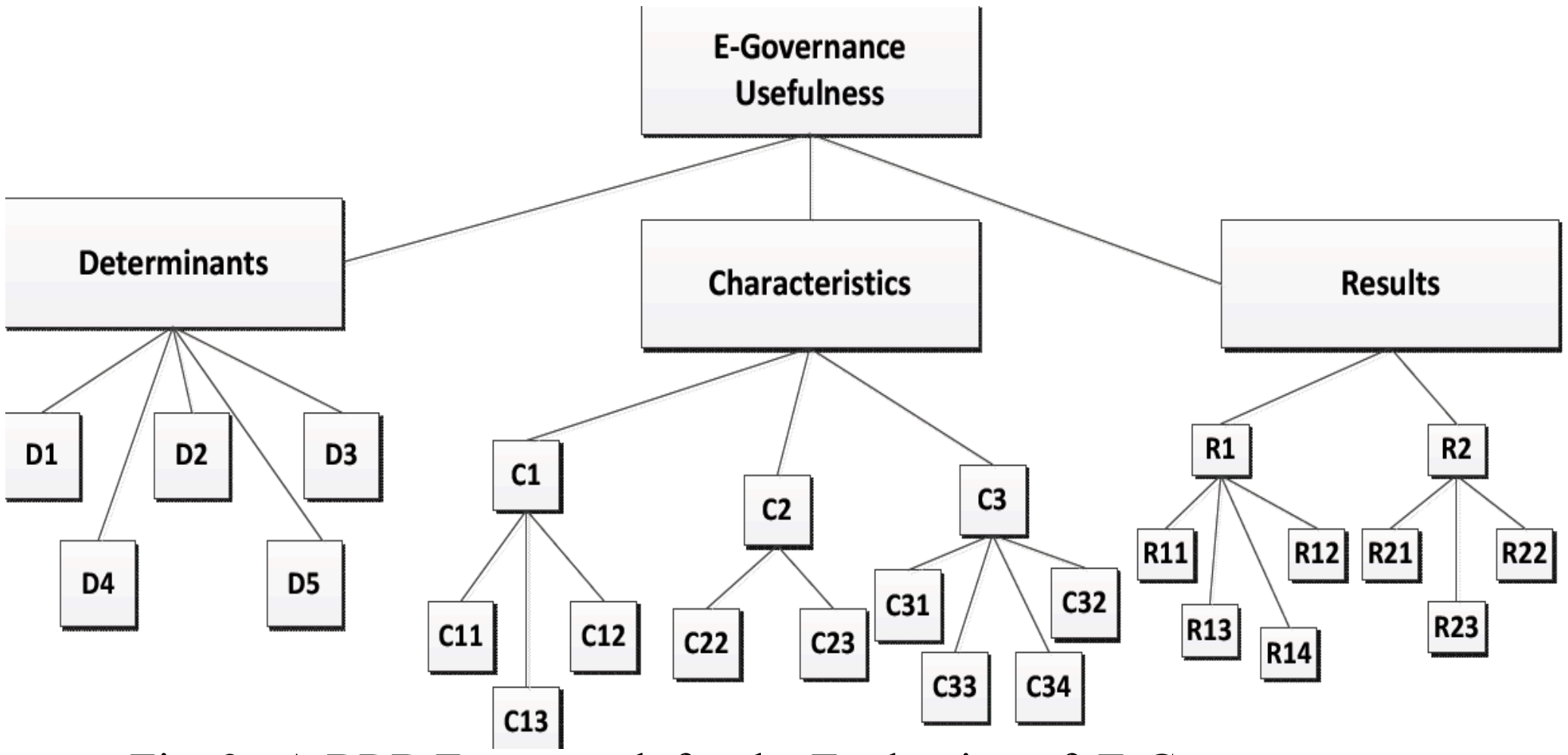

Fig. 3. A BRB Framework for the Evaluation of E-Government

This BRB consists of nine sub-rule-bases, i.e. Determinants, User Environment (C1), Resource Management (C2), Authentication Protocol (C3), 'Result Analysis (R1), 'Result Specification' (R2), 'Characteristics', Results' and 'E-Government Assessment', as illustrated in Table 1. As the rules are exponentially proportional to the referential values and sub-nodes, this evaluation system consists of 4925 rules, making it relatively complex in terms of computation.

A BRB can be established in four ways (D.-L. Xu et al., 2007): by 1) extracting belief rules from expert knowledge, 2) extracting belief rules from the examination of historical data, 3) using the previous rule bases if available, and by 4) using random rules without any pre-knowledge. In this research, the initial BRB was constructed by taking account of domain expert knowledge. The expert assigned equal rule weights to all belief rules i..e "1"; all antecedent attributes assigned also to equal weight i.e "1".

**Table 2**

Determinants′ Sub Rule Base

| Rule ID | Rule Weight | IF | | | | | THEN | | | | |
|---|---|---|---|---|---|---|---|---|---|---|---|
| | | D1 | D2 | D3 | D4 | D5 | Determinants | | | | |
| | | | | | | | X1 | X2 | X3 | X4 | X |
| R0 | 1 | X | X1 | X1 | X1 | X1 | 1 | 0 | 0 | 0 | 0 |
| R1 | 1 | X | X1 | X1 | X1 | X | 0.6 | 0.4 | 0 | 0 | 0 |
| R2 | 1 | X | X1 | X | X | X | 0 | 0.667 | 0.333 | 0 | 0 |
| R3 | 1 | X | X1 | X | X | X | 0 | 0.533 | 0.466 | 0 | 0 |
| R4 | 1 | X | X1 | X | X | X | 0 | 0.4 | 0.6 | 0 | 0 |
| R5 | 1 | X | X1 | X | X | X | 0 | 0.266 | 0.733 | 0 | 0 |
| R6 | 1 | X | X2 | X | X | X | 0 | 0.933 | 0.066 | 0 | 0 |
| R7 | 1 | X | X | X | X | X | 0 | 0 | 0.333 | 0.66 | 0 |
| R8 | 1 | X | X | X | X | X | 0 | 0 | 0.006 | 0.99 | 0 |
| … | … | . | .. | .. | .. | .. | . | .. | … | .. | .. |
| R32 | 1 | X | X | X | X | X | 0 | 0 | 0 | 0.8 | 0. |

*X1 = Excellent, X2 = Very Good, X3 = Good, X4 = Satisfactory, X5 =Poor, D1 = Quality of Information, D2 = Technological Infrastructure, D3 = Organizational Characteristics, D4 = Existing Legal Framework, D5 = Potential Demand*

An example of a belief rule taken from Table 2 is illustrated below.

R2: ***IF*** D1 is X1 ∧ D2 is X1∧ D3 is X2 ∧ D4 is X3 ∧ D5 is X3,***THEN*** Determinants {X1(0.667), X2 (0.333), X3 (0.00), X4 (0.00), X5(0.00)}

In the above belief rule, the belief degrees are attached to the five referential values of the consequent attribute. It is complete since the value summation of degrees of belief is 1. In traditional IF-THEN rules, the consequent is either completely true or completely false. Hence, when such a rule base us used, real world knowledge cannot be represented. In addition, continuous causal relationships between antecedents and consequents cannot be captured in the traditional IF-THEN rule. However, the belief structure in the belief rule base provides more flexibility in representing knowledge of different structures and degrees of complexity, such as continuous and uncertain relationships between antecedents and consequents. For example, in the above rule, the causal relationship between Determinants (consequent) with five antecedents is non-linear, uncertain and complex since their relationship is not proportional in the real world context.

*4.3. BRBES Graphical User Interface*

A system interface can be defined as the media, enabling the interaction between the users and the system. Fig. 4 illustrates a simple interface for the BRBES. This interface facilitates the acquiring of the leaf nodes (antecedent attributes) data of the BRB framework (Fig. 3), which are collected from ICT personnel or the people involved with the project. The system interface enables the displaying of the evaluation results (the top node) and sub-results. For example, Fig. 4 illustrates the result for the data of leaf nodes (D1 = 4, D2 = 4, D3 = 4, D4 = 3, D5 = 4) associated with the 'Determinants' sub-rule-base. From Fig. 4 it can be observed that the degree of belief obtained for the referential values of the consequent attribute "Determinants" of this sub-rule-base is {Excellent (0), Very Good (0), Good (0.1303), Satisfactory (0.8449), Poor (0.0248)}. This was obtained by applying equations (10) and (11). Similarly, the degree of belief of the referential values of the consequent (User Environment) of C1 sub-rule-base is {Excellent (0), Very Good (0.4828), Good (0.5192), Satisfactory (0), Poor (0)} for the input value of the leaf node values (C11-C13) supplied by the ICT personnel. The degree of belief of the referential values of the consequent (Resource Management) of C2 sub-rule-base is {Excellent (0), Very Good (0.2118), Good (0.647), Satisfactory (0.1412), Poor (0)} for the input value of the leaf node, as shown in Fig. 4.

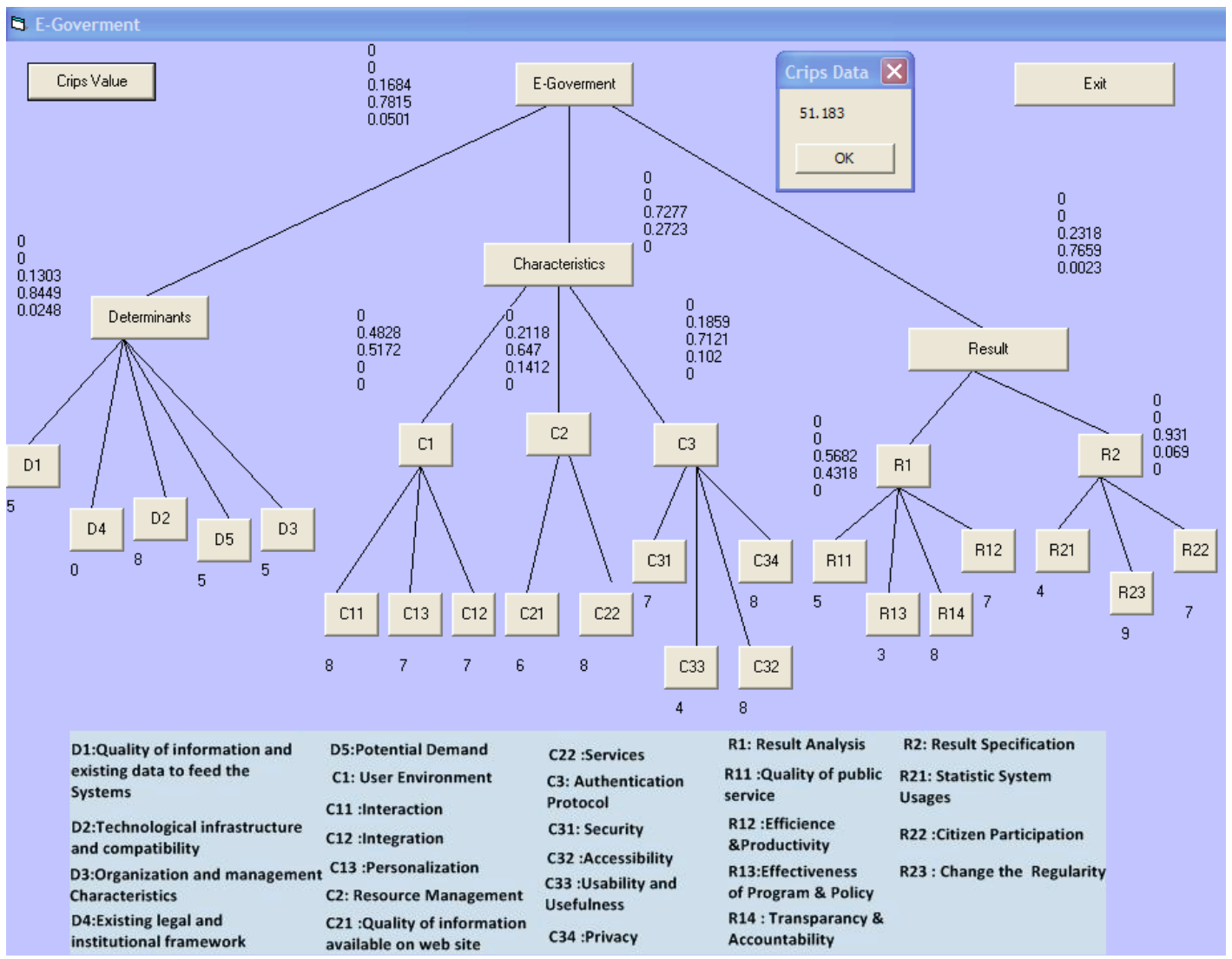

Fig. 4. Graphical User Interface of BRBES

It is interesting to note that the child nodes (C1, C2, C3) of the 'Characteristics' node are not the leaf nodes, and hence, their data cannot be acquired externally to feed the system. These child nodes are actually the consequents of the C1, C2 and C3 sub-rule-bases, and their referential values were already been calculated by the system as the degree of belief. However, in order to obtain the single data value, each referential value of the consequent was multiplied by the utility values, as mentioned in Section 3. The calculated single data value of C1, C2 and C3 were considered to be the antecedent value of the 'Characteristics' sub-rule-base.

Fig. 4 also illustrates the overall e-government assessment (can be considered at the high or aggregated level), which is {Excellent (0), Very Good (0), Good (0.1684), Satisfactory (0.7815), Poor (0.0501)}. This is transformed into crisp value by using equation (12), which is 51.183%, as shown in Fig. 4.

## 5. Application of BRBES and its Validation

In this research, a case study within local e-government in Bangladesh was considered in which the applicability of the BRBES is demonstrated. The way in which users' detailed perceptions of an e-government system can be aggregated to an overall perception of its usefulness and be visualized has already been demonstrated in Fig. 4. This is essential for the simulation of citizen perceptions to be used for design support and decision support. BRBES's ability to incorporate uncertainties will lead to a more powerful modelling than the original LRF model (Luna-Reyes et al., 2012), which uses a simple average of all variables. The BRBES model has other potential uses as well, which we will discuss below.

The geographical units of the country are divided into seven administrative units, called *Divisions*, which are further sub-divided into 64 *districts.* The citizens in three of these districts were considered as the population for this study. The local websites of three of these districts (*Chittagong, Cox's Bazar*, & *Bandarban*) for G2C services were selected as a case for the evaluation of the merits of the system. A detailed description of the services evaluated is beyond the scope of this paper, but the system enables citizens to acquire information on rules and regulations such as the country's present ICT act 'The Right to Information Act' of Bangladesh and its statutes etc. These systems also allow citizens to fill out and submit forms for various purposes, such as land registration and passport application, and in some cases they are able to track the progress online. We compare the BRBES evaluation against the evaluation process of Luna-Reyes et al. framework (2012). We will denote the latter by 'LRF' below. This section presents the survey data collection procedures, and based on this dat,a the validation of the BRBES is also presented by using Receiver Operating Characteristics (ROC) curves.

### *5.1. Survey Data Collection Procedures*

A *multi-staged stratified sampling* technique has been employed in this research (Babbie, 1995). The districts of interest were divided into areas to ensure a precise sampling.

**Table 3**
Sampling Frame for Internal Experts/Personnel

| Categories | Chittagong | Population | Subjects/sample |
|---|---|---|---|
| DC | Chittagong, Cox's Bazar, Bandarban | 11 | 3 |
| ADC | | 25 | 3 |
| Programmer of DC office | | 11 | 3 |
| SP | | 11 | 3 |
| Programmer of SP office | | 11 | 3 |
| CO: from 7 UP such as Cauchua, | Patiya (sub | 80 | 7 |
| **Total** | | **149** | **22** |

*DC: Deputy Commissioner, ADC: Additional Deputy Commissioner, SP: Superintendent of Police, UP: Union Parishad (administrative unit under police station), CO: Computer Operator*

The authors collected data from internal and external users or experts (people who were nominated by the Bangladesh government to execute the E-Government system with proper knowledge and training, and who are considered as internal personnel, whereas the people who are receiving benefits or services from the E-Government System are called external users) through survey questionnaires which were quantitative in nature. In this case study, the survey data were retrieved by structured interviews conducted jointly by the two last authors of the paper.

Table 3 shows the sampling frame which was used for the internal personnel or experts. Furthermore, data from 454 internal and external respondents were collected for the analysis. The sample can be considered sufficient, because sample sizes of more than 30 and less than 500 are appropriate for most research (Roscoe, 1975). This second category will also help to illustrate how BRBES can process data received from multiple stakeholders.

### *5.2. System Validation Using ROC Curves*

Assessment of predictive accuracy is a crucial aspect of evaluating and comparing models, algorithms or technologies that produce the prediction. The ROC curves provide a comprehensive and visually attractive method of summarizing the accuracy of predictions (Gönen, 2007). Therefore, ROC curves have become the standard tool for this purpose, and their use is becoming increasingly common in fields such as finance, clinical applications, atmospheric science, machine learning and many others. Therefore, in this research, ROC curves were used to measure the accuracy of the e-government evaluation generated by the prototype BRBES. This will also allow comparison between the LRF method of evaluating e-government and the prototype BRBES in terms of accuracy. Usually, the accuracy of the results can be measured by calculating the size of the Area Under Curve (AUC) (Gagnon & Peterson, 1998). The larger the area, the higher is the accuracy of the results.

The collected data from the internal and external users/experts associated with ICT projects of the three districts of Bangladesh were used as input data in the prototype BRBES to evaluate the e-government performance. Fig. 5 illustrates the comparisons between ROC curves of BRBES and LRF computations for the 'Determinants' [5(a)], 'Characteristics' [5(b)]

**Table 4**

Comparison of BRBES and LRF

| Method | Determinants | Characteristics | Results | Overall E-Government |
|---|---|---|---|---|
| BRBES | 0.690<br>CI (0.408-0.804) | 0.629<br>CI ( 0.450-0.824) | 0.750<br>CI( 0.432-0.790 ) | 0.725<br>CI (0.501-0.86) |
| LRF | 0. 515<br>CI ( 0.378-0.645) | 0.559<br>CI (0.397-0.693) | 0.5625<br>CI (0.320-0.607) | 0.535<br>CI (0.368-0.612) |

CI = 95% Confidence Interval

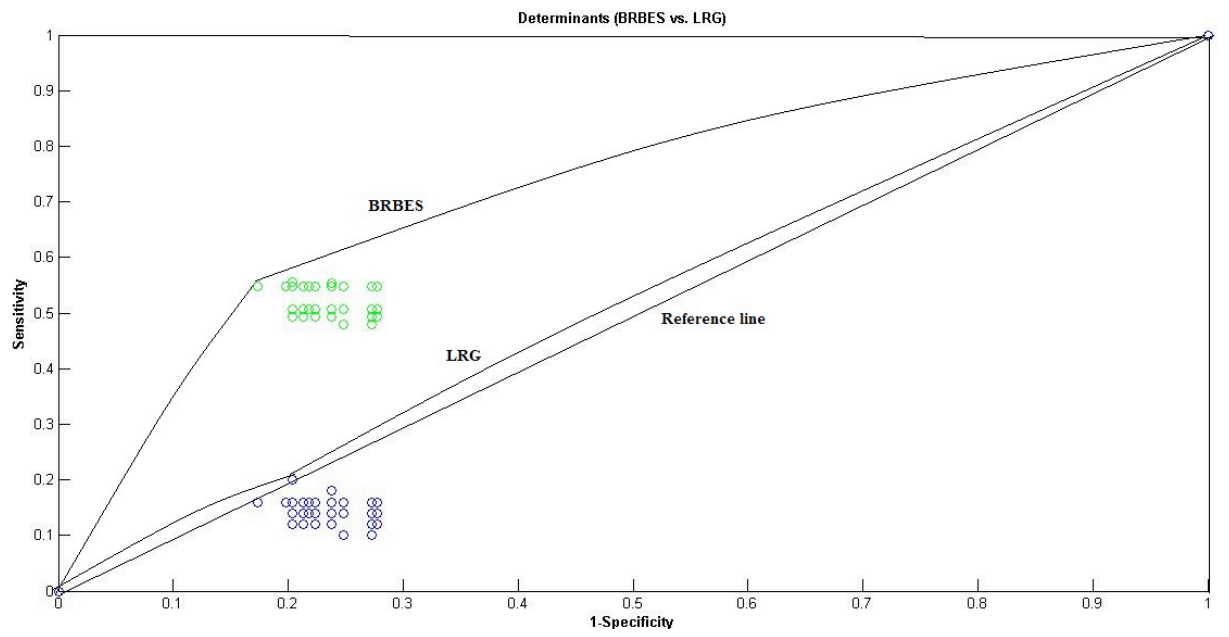

(a)

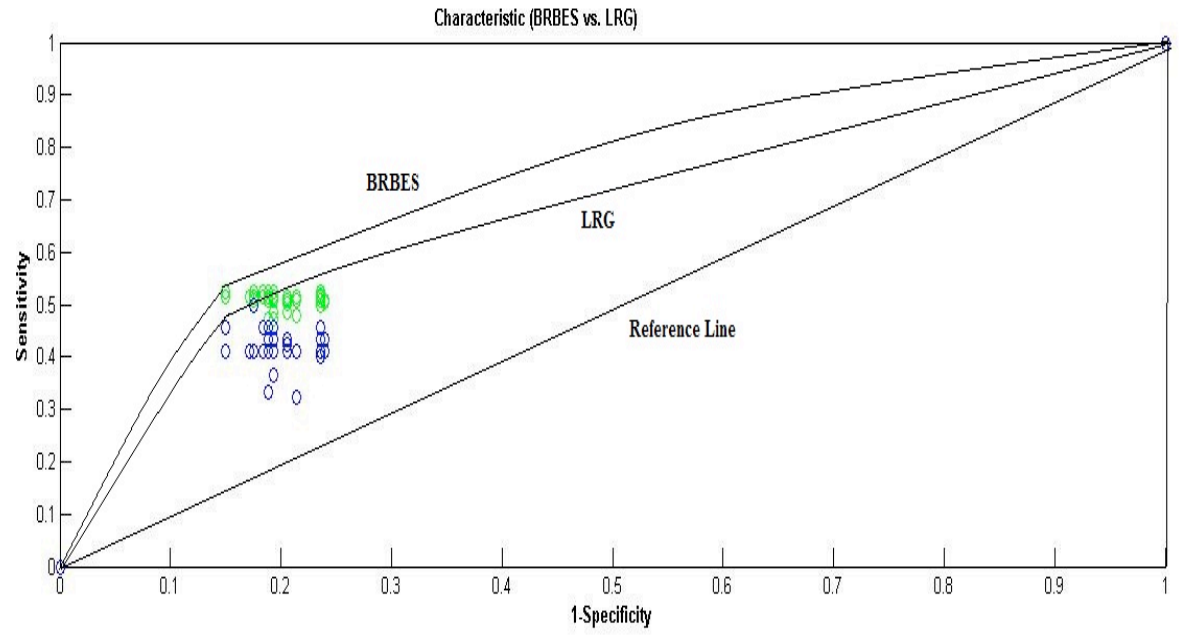

(b)

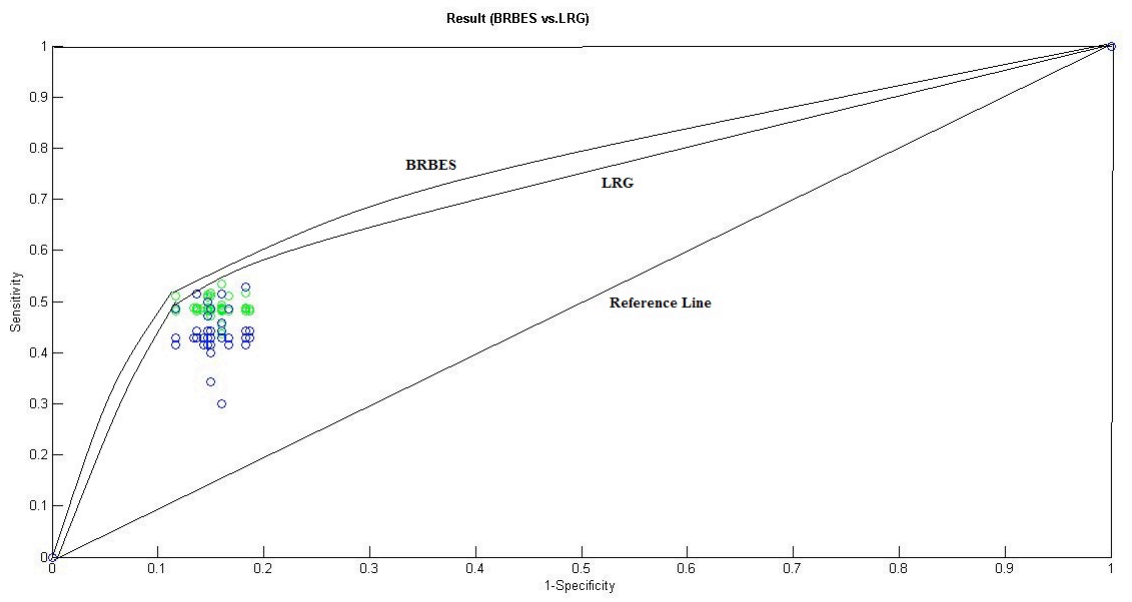

(c)

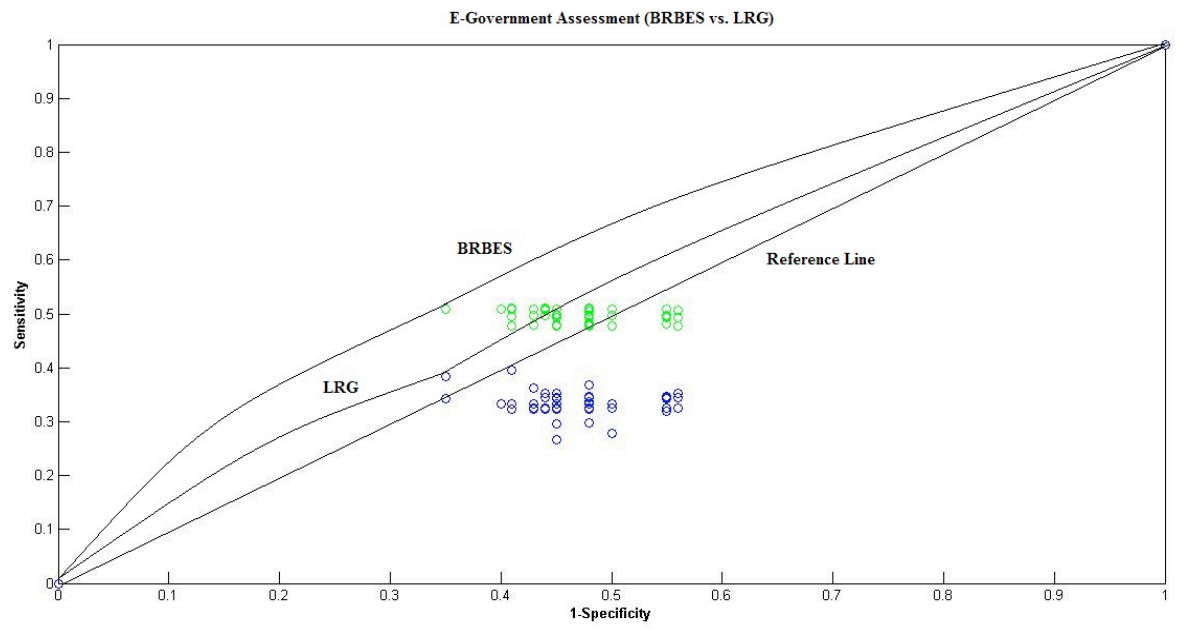

(d)

Fig. 5 Comparison of ROC Curve between BRB Expert System and LRF Method for Characteristics (a), Determinants (b), Results (c) and Over-all E-Government (d)

and 'Results' [5(c)] dimensions of the e-government framework and also of the overall E-Government [5(d)]. We have considered users' perception as the baseline for the comparison between BRBES and LRF, since the expert's opinion on the usefulness on an e-government system in itself is really just an evaluation of how users use the system for their own goals.

From Fig. 5(a), and Table 4, it can be observed that the difference in evaluation of the 'determinant' dimension of the e-government evaluation framework is large since the AUC for BRBES is significantly larger than that of the LRF method. The reason for this is that during our survey, especially external users were unable to evaluate several variables under the different categories, as can be seen from excerpts of the data shown in Tables 5-7. This implies that they have incomplete information or expressed ignorance on these variables. The AUC of BRBES is significantly greater than the LRF method [Table 4 and Fig. 5(a)], because BRBES has the approaches to process this type of uncertainty, whereas the LRF method does not possess this but is based on simple mean value calculation, as mentioned earlier.

Table 5: Determinants data collected during the survey

| ID | OoI | TI | OC | ELF | PD |
|---|---|---|---|---|---|
| 1 | 8 | 0 | 0 | 0 | 0 |
| 2 | 7 | 0 | 0 | 0 | 0 |

Table 6: Characteristics data

| ID | OoI | Ser | IR | IG | PRS | SEC | PRV | ACC | US |
|---|---|---|---|---|---|---|---|---|---|
| 1 | 6 | 6 | 0 | 9 | 6 | 7 | 5 | 0 | 0 |
| 2 | 7 | 5 | 0 | 5 | 7 | 10 | 5 | 0 | 0 |

Table 7: Results data collected during survey

| ID | SU | OPS | EP | EPP | TA | CP | CRF |
|---|---|---|---|---|---|---|---|
| 1 | 8 | 6 | 6 | 7 | 0 | 9 | 0 |
| 2 | 8 | 7 | 2 | 3 | 0 | 4 | 0 |

From Figs. 5(b) and 5(c) as well as on Table 4 it can be inferred that BRBES' generated evaluation results are more accurate than those of the LRF method, since for both of these cases, the AUC of BRBES is greater. It is interesting to note that the LRF method did not consider the top level evaluation of the e-government, but BRBES did consider this phenomenon. The evaluation result of overall e-government is better than that of the LRF method as illustrated in Fig. 5(d) and Table 4. Hence, it can be argued that BRBES' output reflects the user's perception much more than does the LRF method. Several users provided incomplete answers (sometimes due to a lack of knowledge or ignorance of a given issue), and the BRBES takes this into account (see Section 3.2.3). In addition, the BRBES' output would be even closer to the user's perception if the actual weight of each variable was considered during processing of the data, although the system provides the option of changing the weight value of the variable. The weight of variables is assigned “1” (equal weight), as mentioned above.

The area under the ROC curve (AUC) was calculated by using the prism method [Table 4]. This method computes the area under the entire AUC curve by a trapezoidal method, starting at [0,0] and ending at [100,100]. It calculates the area of each trapezoid by calculating the area of the equivalent rectangle. The area under the curve is the sum of areas of all the rectangles (Gagnon & Peterson, 1998). The AUC for BRBES is 0.725 (95% confidence intervals 0.501-0.86), and the AUC of the LRF method is 0.535 (95% confidence intervals 0.368-0.612), as shown in Table 4. The ROC curves of Figure 5 were drawn by developing a function in MatLab. SPSS 11.5 was also used to construct the ROC curves and to calculate the AUC.

## 6. Discussion

The reporting of user perception aggregations demonstrated above is not the pinnacle of functionality in the development of the BRBES. The architecture of the model and system also allows for a large number of adjustments, such as weighting of the nodes, rules and belief degrees, which will increase the accuracy of predictions [see Section 3]. Methods for training this type of expert system already exist (D.-L. Xu et al., 2007). The LRF model can of course also be readjusted to give greater weight to some factors, but its knowledge representation has no inherited support for that. The BRBES can also be used to generate various ‘what if scenarios’ by taking account of different values of the leaf node variables under uncertainty, as mentioned in Sections 3 and 4. This will allow the analysis of an e-government system from various perspectives, which can be used by policy makers to develop appropriate strategies to achieve certain goals. For example, more attention can be paid to less focused factors/variables, shown as leaf nodes in the BRB structure, to upgrade the system's level of performance. This may be necessary to achieve the goal of the e-government system. In addition, the BRBES facilitates the evaluation of the three dimensions, i.e. determinants, characteristics and results in a separate manner, as shown in the previous section. However, simulation is only one of the potential uses of the system.

### *6.1. Prediction of the future perceptions of users*

The present case study tested the correlation between current detailed perceptions and current overall perceived usefulness. However, the BRB model can also be used to infer the future perception of the system, e.g. how current shortcomings of transparency or connectivity will shape the overall perceptions of users. E-government transactions can be considered as an abstract workflow model with input, determinants and results. As such, it can be simulated (see e.g. Tan, Xu, Yang, Xu, & Jiang, 2013). However, while promising, we see this as a non-trivial task due to user heterogeneity, requiring extensive longitudinal data from the problem domain.

### *6.2. Benchmarking*

The BRBES is especially powerful for benchmarking between systems, or versions of the same system, or between administrative units. Ceteri paribus, the system with the higher score is the preferred investment alternative. A system with a relatively low score is a candidate for retirement, upgrading or redesign. BRBES is also useful as a first step in an analysis process in which a subsequent analysis can "zoom in" on explaining the causes for strong or weak performance. Systems with a (relatively) good performance are candidates for "best practice" learning across public sector organisational units.

### *6.3. Automated and Dynamic Composition*

In our proof-of-concept system validation, we have used input from users. In principle, e-government can be automatically reconfigured, so that an evaluation expert system yields a result, and the e-government provider then automatically composes the offer based on that result. The composition can consist of either changing which services that can be accessed by the citizen (e.g. taking down a very unpopular service) or replacing parts in an aggregated service. This presumes that there are several services to choose between, and that there is an overarching system that can choose between components. It also presumes that the information of the users can be collected continuously and automatically. This can be accomplished e.g. through a survey after the use of an e-government service, which is informed by the expert system's variables. BRBES would be able to aquire some of the data automatically (Magoutas & Mentzas, 2010) through web analytics, although the extent to which this can be automated will be dependent on the specific service evaluation and on the context of use. Some interesting work is being done in the area of automated composition through AI generally, also for nondeterministic processes (P. Wang, Ding, Jiang, & Zhou, 2014). Ultimately, the potential for automatic composition of this type is a question of how involved humans could be in the design of systems (Dreyfus, 1992). This debate our present study cannot resolve; we have only shown how it is possible to conduct an automated aggregated evaluation of e-government, despite that the fact that the input may contain uncertainties of various types.

*6.4. The Basis for Stakeholder Negotiations*

Rather than assuming that one stakeholder constitutes the "golden standard", the system can generate a description and a type of summary of the use characteristics of the system, which can form the basis for discussions between stakeholders.

The BRBES describes the aggregate usefulness of a system, and was defined as the way in which the system was able to reach specific goals. Different stakeholders have different goals, and hence they will not be evaluated in the same way. In other words, who should the expert system serve? And if it replaces the expert, who did he or she serve? There are at least two ways to use the system:

1) Refine the weights of the nodes so that it reflects the different goals of the stakeholders. Thus, each stakeholder has an opportunity to manage the complexity of the system, and can further engage in a rational discourse on how to proceed.

2) Reach a consensus among decision-makers before the evaluation (Luna-Reyes et al., 2012) with one least common denominator in terms of goals, and use BRBES to optimize for this goal.

Stakeholder issues ultimately emerge in the introduction of any benchmarking tool and in any evaluation. The same question can also be raised when employing human experts. Who chooses the experts, and whose interests does their knowledge serve? (Galtung, 2003) But this issue is more pertinent in e-government, which also needs to deal with issues such as democracy and transparency. The purpose of our research is not, however, to shift the stakeholder power in public governance in general, even though we feel that it is important not to gloss such issues over. We have only tried to make explicit how the system is able to deal with goals and stakeholders. With BRBES, it is possible to generate various "what if"-scenarios for the consequences for the overall usefulness if a subset of the nodes are improved; these can be used in stakeholder discussions.

*6.5. Limitations*

For some systems and contexts, other baselines may be more appropriate, such as the aggregation of individual items or other data collection techniques. Further case studies that vary as regards context and data collection can corroborate or weaken our findings. Moving the system beyond the field of e-governance would require a completely new data structure and GUI (the tree in Figure 4), but this is beyond the scope of this paper. However, it naturally follows that while. the LRF method, for instance, can be calculated with standard software, the BRRBES will currently need programming expertise and resources spent on development.

The elicitation of knowledge has been direct, in that it has asked for subjective assessment. We could have allocated much more time to eliciting knowledge in different ways (Cooke, 1994), depending on the specific variable. However, we want to bypass detailed technical discussions about each variable through a relatively simple and uniform method of knowledge elicitation for BRBES's, which can show that BRBES can cope with uncertain information, aggregate it into an assessment, and explain the applicability of such expert system assessments.

We believe that ideally, assessments are best made in a process of dialogue with stakeholders, not only because of the possibilities to acquire richer implications for design, but also because sometimes a stakeholder will need a chance to adjust their basis for evaluation. There is no explicit support for this. However, this dialogue is not always an option at every stage of the life cycles of certain systems.

The idea that the user can aggregate complex experience into a simple five-item scale may seem simplistic to the qualitatively minded reader. A related problem is that some e-government systems fulfil many goals, but the aggregation conflates them into one (usefulness). A system may be good for acquiring information about crops, but bad for distance interaction with healthcare, for instance. A less arbitrary test of the overall usefulness of an e-government system may be to ask the users to choose between two alternative systems, or between e-government systems and non-digital government systems. However, we think that the rating will indicate the users' (citizens) current level of satisfaction, and their propensity to criticize governance based on the systems provided. In the same way as the BRBES model assumes that the experience of a system can be aggregated, parliamentary voting is sometimes also the aggregation of the overall life experience into either approval or disapproval of the current government, rather than a completely rational choice between competing alternatives.

The 5-item scale used in our case does not provide many implications for redesign in itself; it is a summative evaluation. It was not designed for inference of concrete design guidance. At best, it can be used for tracing back the roots of bad performance (e.g. weak user interfaces) and to provide some basis for decisions as to what part of the e-government application to improve.

The success of an e-government application is context-dependent, as are so many other IT systems (Beyer & Holtzblatt, 1998). Systems that work well in one domain may fail to do so when transferred. The intention of this evaluation tool, BRBES, was not to discern a context-independent property of the system itself, but rather to measure its usefulness in a given context. Node weighting and variables may need to be adjusted in other contexts.

# 7. Conclusions

The development and application of BRBES to assess e-government systems by using the data collected from both internal personnel and external users of three districts in Bangladesh on 21 variables have been presented. The expert system has implemented a novel methodology known as RIMER, and allows the handling of various types of uncertainty. The data contained numerous instances of data that can be attributed to incompleteness or ignorance, and several of the variables dealt with issues of which users had ambiguous or vague perceptions. The BRBES dealt relatively successfully with this, and hence, it can be considered as a robust tool that can be utilized in e-government assessment. Expert systems of this kind can be trained in order to be used for simulation and predictive modelling. The systems can also be used for more descriptive purposes. BRBES facilitates continuous assessment of e-government projects and allows the identification of the variables that are not contributing. This enables the development of appropriate e-government policies that address the issues associated with a given 'problematic' variable. An example of such a variable may be the quality of information. The BRBES has the scope being used to evaluate e-government projects at various administrative levels in Bangladesh. It can also be used to evaluate the performance of ICT projects in other sectors of the government, such as in education, agriculture and health. This tool (BRBES) can also be used to evaluate the e-government system of other countries in the world.

## References

Accenture. (2004). eGovernment leadership: High performance, maximum value.

Al-Sebie, M., & Irani, Z. (2005). Technical and organisational challenges facing transactional e-government systems: an empirical study. *Electronic Government, an International Journal*, *2*(3), 247–276.

Angulo, C., Cabestany, J., Rodríguez, P., Batlle, M., González, A., & de Campos, S. (2012). Fuzzy expert system for the detection of episodes of poor water quality through continuous measurement. *Expert Systems with Applications*, *39*(1), 1011–1020. doi:10.1016/j.eswa.2011.07.102

Babbie, E. R. (1995). *The Practice of Social Research* (7th ed.). London: Wedsworth Publishing.

Berntzen, L., & Olsen, M. G. (2009). Benchmarking e-Government - A Comparative Review of Three International Benchmarking Studies (pp. 77–82). IEEE. doi:10.1109/ICDS.2009.55

Beyer, H., & Holtzblatt, K. (1998). *Contextual design : defining customer-centered systems*. San Francisco, Calif.: Morgan Kaufmann.

Brown, M. M. (2001). The benefits and costs of information technology innovations: An empirical Assessment of a local government agency. *Pubic Performance & Management Review*, *24*(4), 351 – 366.

Brown, M. M., & Brudney, J. L. (2003). Learning organizations in the public sector? A study of police agencies employing information and technology to advance knowledge. *Public Administration Review*, *63*(1), 30–43.

Cooke, N. J. (1994). Varieties of knowledge elicitation techniques. *International Journal of Human-Computer Studies*, *41*(6), 801–849. doi:10.1006/ijhc.1994.1083

Dawes, S. S. (2008). The evolution and continuing challenges of e-governance. *Public Administration Review*, *68*(s1), S86–S102.

Dreyfus, H. L. (1992). *What computers still can't do: A critique of artificial reason*. Cambridge, Mass: MIT Press.

Esteves, J., & Joseph, R. C. (2008). A comprehensive framework for the assessment of eGovernment projects. *Government Information Quarterly*, *25*(1), 118–132. doi:10.1016/j.giq.2007.04.009

Fang, S., Xu, L., Pei, H., Liu, Y., Liu, Z., Zhu, Y., … Zhang, H. (2014). An Integrated Approach to Snowmelt Flood Forecasting in Water Resource Management. *IEEE Transactions on Industrial Informatics*, *10*(1), 548–558. doi:10.1109/TII.2013.2257807

Gagnon, R. C., & Peterson, J. J. (1998). Estimation of confidence intervals for area under the curve from destructively obtained pharmacokinetic data. *Journal of Pharmacokinetics and Biopharmaceutics*, *26*(1), 87–102.

Galtung, J. (2003). What did the experts predict? *Futures*, *35*(2), 123–145. doi:10.1016/S0016-3287(02)00023-X

Gant, D. B., Gant, J. P., & Jhonson, C. L. (2002). *State web portals: Delivering and financing e-service*. Arlington, VA: The Pricewaterhouse Coopers Endowment for the Business of Government.

Gönen, M. (2007). *Analyzing Receiver Operating Characteristic Curves With SAS*. Sas Inst.

Gupta, M. P., & Jana, D. (2003). E-government evaluation: a framework and case study. *Government Information Quarterly*, *20*(4), 365–387. doi:10.1016/j.giq.2003.08.002

Helbig, N., Ramón Gil-García, J., & Ferro, E. (2009). Understanding the complexity of electronic government: Implications from the digital divide literature. *Government Information Quarterly*, *26*(1), 89–97. doi:10.1016/j.giq.2008.05.004

Hossain, M. S., Chowdhury, M. S. U., & Sarker, S. (2013). Intelligent Tender Evaluation System Using Evidential Reasoning Approach. *International Journal of Computer Applications, New York, USA*, *61*(15), 38–43.

Hossain, S., M., Salah Uddin Chowdury, M., & Sarker, S. (2013). An Intelligent Tender Evaluation System using Evidential Reasoning Approach. *International Journal of Computer Applications*, *61*(15), 38–43. doi:10.5120/10008-4874

Jian-Bo Yang, Jun Liu, Jin Wang, How-Sing Sii, & Hong-Wei Wang. (2006). Belief rule-base inference methodology using the evidential reasoning Approach-RIMER. *IEEE Transactions on Systems, Man, and Cybernetics - Part A: Systems and Humans*, *36*(2), 266–285. doi:10.1109/TSMCA.2005.851270

Karunasena, K., & Deng, H. (2012). Critical factors for evaluating the public value of e-government in Sri Lanka. *Government Information Quarterly*, *29*(1), 76–84. doi:10.1016/j.giq.2011.04.005

Liu, T. I., Singonahalli, J. H., & Iyer, N. R. (1996). DETECTION OF ROLLER BEARING DEFECTS USING EXPERT SYSTEM AND FUZZY LOGIC. *Mechanical Systems and Signal Processing*, *10*(5), 595–614.

Luna-Reyes, L. F., Gil-Garcia, J. R., & Romero, G. (2012). Towards a multidimensional model for evaluating electronic government: Proposing a more comprehensive and integrative perspective. *Government Information Quarterly*, *29*(3), 324–334. doi:10.1016/j.giq.2012.03.001

Magoutas, B., & Mentzas, G. (2010). SALT: A semantic adaptive framework for monitoring citizen satisfaction from e-government services. *Expert Systems with Applications*, *37*(6), 4292–4300. doi:10.1016/j.eswa.2009.11.071

Mahmud, T., & Hossain, M. S. (2012). An Evidential Reasoning-based Decision Support System to Support House Hunting. *International Journal of Computer Applications, New York, USA*, *57*(21), 51–58.

Nielsen, J. (1993). *Usability engineering*. Boston, Mass.: Academic Press.

Raus, M., Liu, J., & Kipp, A. (2010). Evaluating IT innovations in a business-to-government context: A framework and its applications. *Government Information Quarterly*, *27*(2), 122–133. doi:10.1016/j.giq.2009.04.007

Roscoe, J. T. (1975). *Fundamental research statistics for the behavioural sciences* (2nd ed.). New York: Holt, Rinehart and Winston.

Russel, S. J., & Peter, N. (2009). *Artificial Intelligence: A Modern Approach* (3rd ed.). Upper Saddle River, New Jersey: Prentice Hall.

Shan, X., Xin, T., Wang, L., Li, Y., & Li, L. (2013). Identifying Influential Factors of Knowledge Sharing in Emergency Events: A Virtual Community Perspective. *Systems Research and Behavioral Science*, *30*, 367–382.

Stowers, G. N. (2004). *Measuring the performance of e-government*. Washington DC: IBM Center for the Business of Government.

Sun, B., Xu, L., Pei, X., & Li, H. (2003). Scenario-based knowledge representation in case-based reasoning systems. *Expert Systems*, *20*(2), 92–99. doi:10.1111/1468-0394.00230

Sun, R. (1995). Robust reasoning: integrating rule-based and similarity-based reasoning. *Artificial Intelligence*, *75*(2), 241–295. doi:10.1016/0004-3702(94)00028-Y

Tang, D., Yang, J.-B., Bamford, D., Xu, D.-L., Waugh, M., Bamford, J., & Zhang, S. (2012). The evidential reasoning approach for risk management in large enterprises. *International Journal of Uncertainty, Fuzziness and Knowledge-Based Systems*, *20*(supp01), 17–30. doi:10.1142/S0218488512400028

Tan, W., Xu, W., Yang, F., Xu, L., & Jiang, C. (2013). A framework for service enterprise workflow simulation with multi-agents cooperation. *Enterprise Information Systems*, *7*(4), 523–542.

Verdegem, P., & Verleye, G. (2009). User-centered E-Government in practice: A comprehensive model for measuring user satisfaction. *Government Information Quarterly*, *26*(3), 487–497. doi:10.1016/j.giq.2009.03.005

Wang, P., Ding, Z., Jiang, C., & Zhou, M. (2014). Automated web service composition supporting conditional branch structures. *Enterprise Information Systems*, *8*(1), 121–146. doi:10.1080/17517575.2011.584132

Wang, Y.-M., Yang, J.-B., & Xu, D.-L. (2006). Environmental impact assessment using the evidential reasoning approach. *European Journal of Operational Research*, *174*(3), 1885–1913. doi:10.1016/j.ejor.2004.09.059

Xu, D.-L., Liu, J., Yang, J.-B., Liu, G.-P., Wang, J., Jenkinson, I., & Ren, J. (2007). Inference and learning methodology of belief-rule-based expert system for pipeline leak detection. *Expert Systems with Applications*, *32*(1), 103–113. doi:10.1016/j.eswa.2005.11.015

Xu, L., Liang, N., & Gao, Q. (2008). An Integrated Approach for Agricultural Ecosystem Management. *IEEE Transactions on Systems, Man, and Cybernetics, Part C (Applications and Reviews)*, *38*(4), 590–599. doi:10.1109/TSMCC.2007.913894

Yang, H.-H., Liu, J.-L., Chang, M. C. S., & Yang, J.-C. (2012). Improvement of e-government service process via a grey relation agent mechanism. *Expert Systems with Applications*, *39*(10), 9755–9763. doi:10.1016/j.eswa.2012.02.180

Yang, J.-B., & Sen, P. (1997). Multiple Attribute Design Evaluation of Complex Engineering Products Using the Evidential Reasoning Approach. *Journal of Engineering Design*, *8*(3), 211–230. doi:10.1080/09544829708907962

Yang, J.-B., & Singh, M. G. (1994). An evidential reasoning approach for multiple-attribute decision making with uncertainty. *Systems, Man and Cybernetics, IEEE Transactions on*, *24*(1), 1–18.

Yang, Jian-Bo, & Singh, M. G. (1994). An evidential reasoning approach for multiple-attribute decision making with uncertainty. *IEEE Transactions on Systems, Man, and Cybernetics*, *24*(1), 1–18. doi:10.1109/21.259681

Zhou, Z.-J., Hu, C.-H., Xu, D.-L., Yang, J.-B., & Zhou, D.-H. (2010). New model for system behavior prediction based on belief rule based systems. *Information Sciences*, *180*(24), 4834–4864. doi:10.1016/j.ins.2010.08.016

**Mohammad Shahadat Hossain** is a Professor of Computer Science and Engineering at the Chittagong University (CU), Bangladesh. He did both his MPhil and PhD in Computation from the University of Manchester Institute of Science and Technology (UMIST), UK in 1999 and 2002 respectively. His current research areas include e-government, the modeling of risk and uncertainty using evolutionary computing techniques. Investigation of pragmatic software development tools and methods for information systems in general and for expert systems in particular are also his area of research.

**Pär-Ola Zander** is associate professor at Aalborg University, where he leads the ICT4D research group. He has previously worked at the department of computer science at Aarhus University, and holds a PhD degree in information systems from Lund University, Sweden. His current research interest is user involvement in system development and e-government, in particular in areas pertaining to development research.

**Md. Sarwal Kamal** did his B.Sc. in Computer Science and Engineering from the University of Chittagong. Currently he is pursuing his M.Phil in the Department of Computer Science and Engineering of Chittagong University. He is the lecturer of Computer science and Engineering Department of BGC Trust University, Bangladesh. His research interest includes e-government and evolutionary computing.

**Linkon Chowdhury** did his B.Sc. in Computer Science and Engineering from the University of Chittagong. Currently he is pursuing his M.Phil in the Department of Computer Science and Engineering of Chittagong University. He is the lecturer of Computer science and Engineering Department of BGC Trust University, Bangladesh. His research interest includes e-government, evolutionary computing, geographical information systems and software development.